\documentclass{article}

\PassOptionsToPackage{numbers, sort&compress}{natbib}
\usepackage{arxiv}
\RequirePackage{natbib}
\usepackage{microtype}
\usepackage{graphicx}
\usepackage{booktabs} %
\usepackage{paralist}
\usepackage{enumitem}
\usepackage{adjustbox}
\usepackage{sidecap}
\usepackage{enumitem}
\sidecaptionvpos{figure}{t}

\usepackage{wrapfig}
\usepackage{todonotes}

\usepackage{caption}
\usepackage{subcaption}

\usepackage[ruled,vlined]{algorithm2e}

\usepackage{import}

\usepackage[utf8]{inputenc} %
\usepackage[T1]{fontenc}    %
\usepackage{url}

\usepackage{amsmath,amsfonts,bm}

\def\eqref#1{equation~\ref{#1}}

\def\1{\bm{1}}

\DeclareMathAlphabet{\mathsfit}{\encodingdefault}{\sfdefault}{m}{sl}
\SetMathAlphabet{\mathsfit}{bold}{\encodingdefault}{\sfdefault}{bx}{n}

\usepackage{amsthm}
\usepackage{amsfonts, amssymb}       %
\usepackage{nicefrac}       %
\usepackage{microtype}      %
\usepackage{amsmath}
\usepackage{bbm}
\usepackage{bibentry}
\usepackage{float}
\usepackage{dsfont}
\usepackage{setspace}
\usepackage{todonotes}
\usepackage{wrapfig}
\usepackage{algorithmic}
\usepackage{multicol}
\usepackage{multirow}
\usepackage{cleveref}
\nobibliography*

\title{Meta-PDE:\\Learning to Solve PDEs Quickly Without a Mesh}

\author{%
  Tian Qin \\
  School of Engineering and Applied Sciences\\
  Harvard University\\
  \texttt{tqin@g.harvard.edu} \\
  \And
  Alex Beatson \\
  Redesign Science \\
  \texttt{alex@redesignscience.com} \\
  \And
  Deniz Oktay \\
  Department of Computer Science \\
  Princeton University \\
  \texttt{doktay@princeton.edu} \\
  \And
  Nick McGreivy \\
  Department of Astrophysical Sciences \\
  Princeton University \\
  \texttt{mcgreivy@princeton.edu} \\
  \And
  Ryan P. Adams \\
  Department of Computer Science \\
  Princeton University \\
  \texttt{rpa@princeton.edu} \\
}

\begin{document}

\maketitle

\begin{abstract}
\looseness=-1 Partial differential equations (PDEs) are often computationally challenging to solve, and in many settings many related PDEs must be be solved either at every timestep or for a variety of candidate boundary conditions, parameters, or geometric domains.
We present a meta-learning based method which learns to rapidly solve problems from a distribution of related PDEs.
We use meta-learning (MAML and LEAP) to identify initializations for a neural network representation of the PDE solution such that a residual of the PDE can be quickly minimized on a novel task. 
We apply our meta-solving approach to a nonlinear Poisson's equation, 1D Burgers'
equation, and hyperelasticity equations with varying parameters, geometries, and boundary conditions.
The resulting Meta-PDE method finds qualitatively accurate solutions to most problems within a few gradient steps; for the nonlinear Poisson and hyper-elasticity equation this results in an intermediate accuracy approximation up to an order of magnitude faster than a baseline finite element analysis (FEA) solver with equivalent accuracy.
In comparison to other learned solvers and surrogate models, this meta-learning approach can be trained without supervision from expensive ground-truth data, does not require a mesh, and can even be used when the geometry and topology varies between tasks.

\end{abstract}
\section{Introduction}
\label{sec:metapde-intro}
Partial differential equations (PDEs) can be used to model many physical, biological, and mathematical systems. Such systems include those governing thermodynamics, continuum mechanics, and electromagnetism. Applications of PDEs outside science include modeling of traffic, populations, optimality of continuous control, and finance.
Analytical solutions are rarely available for PDEs of practical importance; thus, computational methods must be used to find approximate solutions.

\looseness=-1 One of the most widely used approximation methods is finite element analysis (FEA).
In FEA, the continuous problem is discretized and the solution is represented by a piecewise polynomial on a mesh.
Solving PDEs with FEA can be computationally prohibitive, particularly when the problem geometry requires a fine mesh; the size of the system to be solved grows in proportion to the number of mesh cells.
The main purpose of this paper is to use gradient-based meta-learning to accelerate solving PDEs with physics-informed neural networks (PINNs).
This results in solvers that can achieve accurate solutions at reduced computational cost relative to FEA.
Although these solvers have an initial training cost, they may provide computational savings in problems where a PDE must be solved repeatedly.
Such problems could include parameter identification, design optimization, or in the solution of coupled time-dependent PDEs where, e.g., an elliptic equation is solved at each timestep.

PINNs use a neural network (NN) to represent the approximate solution of a PDE.
The idea was popularized by \citet{raissi2019physics}, and has been widely researched since.
The key advantage of PINNs over traditional numerical solvers is that the PINN is able to provide a solution without the need to discretize the problem domain, i.e., the learned PDE solution is mesh-free.
However, PINNs suffer from two major issues that limit their utility as forward solvers~\citep{perdikaris2022respectingcausality}.
First, PINNs have not demonstrated the ability to solve all PDEs. In particular, PINNs tend to struggle to solve time-dependent PDEs whose solutions exhibit chaotic behavior or turbulent flow~\citep{causalitypinn}.
Second, vanilla PINNs tend to be dramatically slower than classic numerical methods.
We attempt to mitigate this second issue by applying meta-learning to partially amortize the cost of optimization, thereby reducing the time required to find an accurate solution on a particular problem.

Forward-solving with PINNs requires optimization (i.e., learning); thus to accelerate forward solving we need to accelerate learning. 
Recent work in meta-learning has focused on how to construct learning algorithms that can adapt to a new task with as little additional training as possible.
We focus on gradient-based meta learning algorithms such as MAML~\citep{finn2017model}, REPTILE~\citep{nichol2018first}, and LEAP~\citep{flennerhag2018transferring}.
These algorithms view meta-learning as a bi-level optimization problem: the inner learning loop optimizes the model parameters for a given task, and the outer learning loop optimizes the inner loop's learning process across the tasks that the inner loop might encounter.

\looseness=-1 \textbf{Main contribution: } We introduce a framework which accelerates PDE solving by combining  meta-learning and PINNs: Meta-PDE. PDE solving is accelerated by using gradient-based meta-learning techniques such as LEAP and MAML to train a PINN initialization which will converge quickly when optimized for a task drawn from a set of related tasks.
The distribution of problems consists of different parameterizations of the PDE, such as different boundary conditions, initial conditions, the coefficients in the governing equation, or even the problem domain of the PDE.
During deployment, the meta-learned model can be used to produce fast solutions to instances of PDEs in the distribution.

Our scheme has three important properties:
\begin{enumerate}
    \item Training does not require supervised data provided by PDE solvers. 
This is in contrast to other learned PDE solvers and surrogate models, which typically train the solver to minimize the residual between the ground-truth solution and the predicted solution \cite{data_driven_discretizations,les_closure,beatson2020learning,message_passing_neural_pde_solvers,ml_accelerated_cfd,ml_to_augment_sims,2d_learned_advection}. We instead minimize a residual of the governing equation (see \cref{sec:PINNs}), eliminating the need for ground-truth data.
\item Meta-PDE is mesh-free and can be used on a broad class of boundary value problems, including problems with arbitrary geometries, and both time-dependent and time-independent PDEs. 
\item Geometry, boundary conditions, and even the PDE are free to vary between tasks. Meta-PDE does not require a vector representation of the factor of variation between tasks which can be input to a neural network: instead, the user supplies an appropriate sampler for the domain and a loss function to measure the residual of the PDE solution. This differs from other learned PDE solvers and surrogate models, which are almost always trained for a single mesh or geometry and cannot be used when the geometry varies \cite{chaudhuri2019multigrid,li2020fourier,deepmind_turbulence_sims,learning_neural_pde_solvers}. 
\end{enumerate}

Previous work has also explored meta-learning for PINNs. \citet{de2021hyperpinn} meta-trains a hyper-network that for each task can generate weights for a small neural network. The small neural network then becomes the approximate solution to the PDE.
\citet{psaros2022metalearninglossfunctions} meta-learns a loss function which is used to optimize the NN; this is found to achieve performance benefits in comparison to both hand-crafted and online adaptive loss functions.
\citet{penwarden2021physics} proposes a meta-learning approach which, like our Meta-PDE approach, learns an initialization of weights such that the NN can be optimized quickly. \citet{penwarden2021physics} compares MAML to other meta-learning approaches that initialize weights using a linear combination of basis functions.
An important way in which these approaches are different from our MAML and LEAP-based approach is that the weight initialization depends on the task parameter that is varied.
\citet{penwarden2021physics} finds that MAML achieves poor performance, only marginally better than random initialization; both our approach and the recently published \citet{liu2022novel} (which uses REPTILE instead of MAML) come to the opposite conclusion. 

\newpage
\section{Meta-learning mesh-free PDE operators}
We take our problems to be defined on the spatial domain $\Omega \subset \mathbb{R}^{d}$ with boundary $\partial \Omega$, and consider time-dependent PDEs
\begin{align}
\begin{split}
\frac{\partial}{\partial t}u(\bm{x}, t) + \mathcal{F}\left[ u(\bm{x}, t)\right] &= 0 \textnormal{  \hspace{1.67cm}   } \bm{x} \in \Omega,\textnormal{  \hspace{1.0cm}   } t \in [0, T], \\
\mathcal{G}(u)(\bm{x}, t) &= 0 \textnormal{  \hspace{1.67cm}   } \bm{x} \in \partial\Omega,\textnormal{  \hspace{0.76cm}   } t \in [0, T], \\
u(\bm{x}, 0) &= \bar{u}_0(\bm{x}) \textnormal{  \hspace{1.0cm}   } \bm{x} \in \Omega,
\end{split}
\end{align}
as well as time-independent PDEs which only have spatial dependence:
\begin{align}
\begin{split}
\mathcal{F}\left[ u(\bm{x})\right] &= 0\textnormal{  \hspace{1.0cm}   }\bm{x} \in \Omega, \\
\mathcal{G}(u)(\bm{x}) &= 0\textnormal{  \hspace{1.0cm}   }\bm{x} \in \partial\Omega. 
\end{split}
\end{align}
For time-dependent PDEs the time horizon is $[0, T]$. 
The function $u(\bm{x},t)$ is the (unknown) solution to the PDE, while $\bar{u}_0(\bm{x})$ is the initial condition.
In both cases $\mathcal{F}$ and $\mathcal{G}$ are governing equation and boundary operators that involve $u$ and partial derivatives of $u$ with respect to spatial coordinates $\bm{x}$.

\subsection{Physics-Informed Neural Networks (PINN)\label{sec:PINNs}}
The goal of a PINN is to represent the approximate solution $u(\bm{x},t)$ with a neural network $f_\theta(\bm{x},t)$. 
Doing so requires learning $\theta \in \mathbb{R}^p$ such that $f_\theta$ approximates the solution to the PDE over the problem domain.
Learning these parameters is an optimization problem, which requires defining a loss function whose minimum is the solution of the PDE.
We choose a ``physics-informed loss'' which consists of an integral of the local residual of the differential equation over the problem domain as well as the boundary.
Analytically, the residual should be integrated over the problem domain to compute the residual from satisfying the governing equation, and the initial condition and integrated over the boundary domain to compute the residual from satisfying the boundary conditions.
For time-independent PDEs, this loss function is
\begin{align}\label{eq:pinn_true_loss_function}
  \mathcal{J}(u) &= \int_{\Omega} \left|\left| \mathcal{F}(u)(\bm{x})\right|\right|^2_2 \, dx + \int_{\partial\Omega} \left|\left| \mathcal{G}(u)(\bm{x})\right|\right| _2^2 \, dx.
\end{align}
For time-dependent PDEs, this loss function is: 
\begin{align}\label{eq:pinn_true_loss_function_time}
  \mathcal{J}(u) = & \int_{\Omega} \left|\left| \frac{\partial}{\partial t}u(\bm{x}, t) + \mathcal{F}(u)(\bm{x}, t)\right|\right|^2_2 + \left|\left| u(\bm{x}, 0) - \bar{u}_0(\bm{x}) \right|\right| _2^2\, dx +
   \int_{\partial\Omega} \left|\left| \mathcal{G}(u)(\bm{x}, t)\right|\right| _2^2 \, dx.
\end{align}
During training, we randomly and uniformly sample collocation points from the PDE domain~$\Omega$ and boundary $\partial \Omega$ and use these points $\mathcal{C} \in \Omega$ and~${\partial \mathcal{C} \in \partial \Omega}$ to form a Monte Carlo estimate of the true loss. For time-independent PDEs, this training loss is
\begin{align}\label{eq:pinn_training_loss}
    \mathcal{L}_{\text{PINN}}(f_\theta) = \frac{1}{|\mathcal{C}|} \sum_{\bm{x} \in \mathcal{C}} \left|\left| \mathcal{F} (f_{\theta})(\bm{x}) \right|\right|_2^2  + \frac{1}{|\partial \mathcal{C}|}  \sum_{{\bm{x} \in \mathcal{\partial \mathcal{C}}} } \left|\left|\mathcal{G}(f_{\theta})(\bm{x})\right|\right|_2^2.
\end{align}
For time-dependent PDEs, the training loss is
\begin{align}\label{eq:pinn_training_loss_time}
\begin{split}
    \mathcal{L}_{\text{PINN}}(f_\theta) = & \frac{1}{|\mathcal{C}|} \sum_{\bm{x} \in \mathcal{C}}\left|\left| \frac{\partial}{\partial t}(f_\theta)(\bm{x}, t) + \mathcal{F} (f_{\theta})(\bm{x}, t) \right|\right|_2^2  +\\ 
     & \frac{1}{|\partial \mathcal{C}|}  \sum_{{\bm{x} \in \mathcal{\partial \mathcal{C}}} } \left|\left|\mathcal{G}(f_{\theta})(\bm{x}, t)\right|\right|_2^2 + \frac{1}{|\mathcal{C}|} \sum_{\bm{x} \in \mathcal{C}} \left|\left| f_{\theta}(\bm{x}, 0) - \bar{u}_0(\bm{x}) \right|\right| _2^2.
\end{split}
\end{align}
When the training converges, we expect $f_{\theta}$ to approximately satisfy the above equations, meaning that $\mathcal{L}_{\text{PINN}}(f_\theta)$ should be nearly zero.

\subsection{Meta-PDE}

Meta-PDE involves using gradient-based meta-learning to amortize the training time needed to fit~$f_{\theta}$ on a problem drawn from a distribution of parameterized PDEs. We focus specifically on two meta-learning methods: LEAP~\citep{flennerhag2018transferring} and MAML~\citep{finn2017model}. We describe LEAP-based Meta-PDE briefly here. MAML-based Meta-PDE is a straightforward extension and is described in Appendix \ref{appdx:meta-pde-dea}.

Most PDEs can be fully specified by their domain, boundaries, an operator representing governing equations, and an operator representing boundary conditions. When using Meta-PDE as a surrogate to compute an approximate solution to a given parametrization of the PDE (one task), the inputs to the Meta-PDE model imitate this general specification:
\begin{itemize}
  \item A sampler $s(\Omega)$ which returns points in the domain $\Omega$,
  \item A sampler $s(\partial\Omega)$ which returns points on the boundary $\partial\Omega$,
  \item An operator $\mathcal{F}$ representing governing equations,
  \item An operator $\mathcal{G}$ representing boundary conditions.
\end{itemize}
The operators $\mathcal{F}$ and $\mathcal{G}$ may be supplied directly and do not require a particular parametric form. The geometric dimension $\mathbb{R}^{d_\Omega}$ and solution dimension $\mathbb{R}^{d_u}$ must remain fixed across PDEs in the distribution, even though $\Omega$ is allowed to vary. The samplers and operators are sufficient to construct an estimator $\hat{\mathcal{L}}$ for the task loss $\mathcal{L}$ using Eqn.\ref{eq:pinn_training_loss} for time-independent problems and Eqn.\ref{eq:pinn_training_loss_time} time-dependent problems.
Although $\hat{\mathcal{L}}(f)$ is unbiased as long as $s(\cdot)$ return points with uniform probability over their supports, note that unbiased estimation is not necessarily essential if we are interested in the case where $\mathcal{L}(f) = \hat{\mathcal{L}}(f) = 0$. Biased sampling will not change the minimizer of the energy estimator if we have a sufficiently expressive hypothesis class for $f$.

The LEAP-based Meta-PDE method learns the model initialization $\theta^0 \in \mathbb{R}^p$ for a neural network $f_\theta$, which can then be trained to approximate the solution $u: \mathbb{R}^{d_\Omega} \to \mathbb{R}^{d_u}$ of an individual parametrization of the PDE. To learn $\theta^0$, we start with a distribution of tasks, where each task represents a different parameterization of the PDE. Each task is specified by samplers and constraint operators for the boundary and loss. Then we draw a batch of $n$ tasks with individual loss functions $\hat{\mathcal{L}}_i$, $i \in [n]$. The initialization for each inner task is $\theta^0$, and is updated by the inner gradient update rule. During each inner gradient update, we update the meta-gradient per the LEAP algorithm. We unroll the inner learning loop $K$ steps to find $f_{\theta_i^K}$: the approximate solution for each task $i$ in the batch. After unrolling $K$ update steps for $n$ tasks, we update the learned model initialization $\theta^0$ with the meta-gradient:
\begin{align}
    \theta^0 \leftarrow \theta^0 - \beta \nabla_{\theta^0}\sum_{i=1}^n \frac{1}{n} d(\theta^0; M_i),
\label{eq:leap-update}
\end{align}
where $d(\theta^0; M_i)$ is the distance of the gradient path for task $i$ on its manifold $M_i$, as specified in \citet{flennerhag2018transferring}. MAML involves a slightly different loss function and also learns step sizes for each parameter.

During deployment time, a ``forward pass'' computes an approximate solution for a given PDE parametrization with $K$ steps of stochastic optimization. The $K$ gradient steps minimize the training loss for the task $\mathcal{L}(f)$. If the model has been trained with LEAP-based Meta-PDE method, it will start from the meta-learned model initialization $\theta^0$. If the model has been trained with MAML-based Meta-PDE method, it will start from the meta-learned model initialization $\theta^0$ and the step size $\alpha$ will be also be specified for each parameter and each step: 
\begin{align}
  \theta^k = \theta^{k-1} - \alpha \nabla_{\theta^{k-1}} \hat{\mathcal{L}}(f_{\theta^{k-1}}) \quad k = 1, \ldots, K\,.
\end{align}
In both cases, the Meta-PDE method returns the approximate solution $f_{\theta^K}$, the neural network with the final set of parameters. One could further fine tune the model beyond $K$ gradient steps to achieve higher solution accuracy at the cost of longer solving time.

\newpage
\section{Experiments\label{sec:experiments}}
We evaluate the application of Meta-PDE to three example PDE problems: the nonlinear Poisson's equation, the 1D Burgers’ equation, and the hyper-elasticity equation. 
We discuss the results of training on Burgers' equation in \cref{sec:discussion}, as well as in Appendix \ref{appdx:burgers}.
Meta-PDE methods are implemented in JAX \citep{jax2018github}. 
In training, we sample points uniformly on the domain and the boundary. 
Our LEAP and MAML-based Meta-PDE models use a small NN with sinusoidal activation functions.
The sinusoidal activation is initialized according to the scheme in \citet{sitzmann2020implicit}, although we replace $\omega_0 = 30.0$ in that paper with $\omega_0 = 3.0$ to avoid numerical issues when taking higher-order derivatives of a neural network's input-output function. 
Gradients in both inner and outer loop are clipped to have maximal norm $100.0$. 
Additional NN hyperparameters are in Table \ref{tbl:hparams} while training hyperparameters are in Table \ref{tbl:training}. 
We compare Meta-PDE's performance with a baseline FEniCS~\citep{LoggMardalEtAl2012a,AlnaesBlechta2015a} solver. We use the Mumps linear solver backend. 

During deployment, the Meta-PDE solutions can be further improved by extending the number of ``inner'' training steps beyond what is used at training time. In our MAML-based Meta-PDE method for example, we train the meta-learned initialization using 5 inner-gradient steps. At deployment time, we can refine the Meta-PDE solution by using a greater number of inner-gradient steps. We compare the speed/accuracy trade-off achieved by varying the number of inner-gradient steps Meta-PDE takes during deployment with the speed/accuracy trade-off achieved by varying the resolution of the mesh used in FEA.

\begin{table}[htbp]
\caption{Neural network hyperparameters for our Meta-PDE methods}
\label{tbl:hparams}
\centering
\begin{adjustbox}{max width=\textwidth}
\begin{tabular}{llllllll}
    \toprule
    PDE Problem & Meta-PDE Method & \multicolumn{6}{c}{Hyperparameters}  \\
    &  & Num. of layers & Layer Size & Activation & Inner Steps & Inner LR & Outer LR \\
    \midrule
    Nonlinear Poisson's & \multirow{3}{*}{MAML} & 3 &  \multirow{3}{*}{64} & \multirow{3}{*}{$\sin$} &  \multirow{3}{*}{5} & $1.0\times10^{-4}$ & $1.0\times10^{-5}$  \\
    Burgers' &  & 8 &  &  &  & $1.0\times10^{-4} $ & $1.0\times10^{-5}$ \\
    Hyper-elasticity &  & 5 &  & &  & $1.0\times10^{-5} $ & $5.0\times10^{-6}$  \\
    \bottomrule
    Nonlinear Poisson's & \multirow{3}{*}{LEAP} & 5 & 64 & \multirow{3}{*}{$\sin$} & 60 & $2.5\times10^{-5} $ & $5.0\times10^{-5}$  \\
    Burgers' &  & 10 & 128 & & 80 & $1.0\times10^{-6} $ & $5.0\times10^{-5}$  \\
    Hyper-elasticity &  & 10 & 128 &  & 20 & $5.0\times10^{-6} $ & $5.0\times10^{-6}$  \\
    \bottomrule
  \end{tabular}
\end{adjustbox}
\end{table}
\begin{table}[htbp]
\caption{\small 
Training hyperparameters for our meta-PDE methods and training time on one NVIDIA T4 GPU}
\label{tbl:training}
\centering
\begin{adjustbox}{max width=\textwidth}
\begin{tabular}{llllllll}
    \toprule
    PDE Problem & Meta-PDE Method & \multicolumn{6}{c}{Hyperparameters}  \\
    & & Batch Size & Sampled Points & Iterations & Training Time & Inner Optimizer & Outer Optimizer \\
    \midrule
    Nonlinear Poisson's & \multirow{3}{*}{MAML} & \multirow{3}{*}{8} & 2048 & 120,000 & 4 hrs & \multirow{3}{*}{SGD} & \multirow{3}{*}{Adam} \\
    Burgers' &   &  & 1024 & 60,000 & 11 hrs  &  & \\
    Hyper-elasticity &   &  & 1024 & 180,000 & 21 hrs &  & \\
    \midrule
    Nonlinear Poisson's & \multirow{3}{*}{LEAP}  & \multirow{3}{*}{8} & 4096 & 55,000 & 5 hrs  &  \multirow{3}{*}{Adam} &  \multirow{3}{*}{Adam}\\
    Burgers' &   &  & 2048 & 7,000 & 7 hrs &  & \\
    Hyper-elasticity &   &  & 1024 & 140,000 & 8 hrs &  & \\
    \bottomrule
  \end{tabular}
\end{adjustbox}
\end{table}

\subsection{Nonlinear Poisson's Equation}
Poisson's equation is one of the most ubiquitous equations in physics. 
For example, the linear Poisson equation calculates the electrostatic or gravitational field caused by electric charges or mass particles.
Solving systems of coupled time-dependent PDEs often requires the solution of a Poisson equation at each timestep \cite{orszag1969numerical}.
Since the linear Poisson's equation can be solved analytically, we demonstrate Meta-PDE on a nonlinear Poisson problem with varying source terms, boundary conditions, and geometric domain. This nonlinear Poisson's equation takes the form
\begin{align*}
\nabla \cdot \left[ (1 + 0.1 u^2) \nabla u(\bm{x}) \right]&= f(\bm{x}) \quad &\bm{x} \in \Omega\\
u(\bm{x}) &= b(\bm{x}) \quad &\bm{x} \in \partial\Omega
\end{align*}
where $u \in \mathbb{R}^1$ and $\Omega \subset \mathbb{R}^2$. Using our notation from the previous section, this is equivalent to constraining the solution in the domain with an operator~${\mathcal{F}(u) = ((1 + 0.1 u^2) \nabla u) - f}$, and constraining the solution on the boundary with an operator~${\mathcal{G}(u) = u - b}$.

The domain $\Omega$ is a disc-like shape centered at the origin, defined in polar coordinates with varying radius about the origin 
$r(\theta) = r_0[1 + c_1 \cos(4\theta) + c_2 \cos(8\theta)]$,
where the varying parameters are $c_1, c_2 \sim \mathcal{U}(-0.2, 0.2)$. The source term $f$ is a sum of radial basis functions
$f(\bm{x}) = \sum_{i=1}^3 \beta_i \exp{||\bm{x} - \mu_i||_2^2}$,
where $\beta_i \in \mathbb{R}^1$ and $\mu_i \in \mathbb{R}^2$ are both drawn from standard normal distributions. The boundary condition $b$ is a periodic function, defined in polar coordinates as
$b(x) = b_0 + b_1 \cos(\theta) + b_2 \sin(\theta) + b_3 \cos(\theta) + b_4 \sin(\theta)$,
where the parameters $b_{0:4} \sim \mathcal{U}(-1, 1)$.

\begin{figure}
     \centering
     \begin{subfigure}{0.485\textwidth}
         \includegraphics[width=\textwidth]{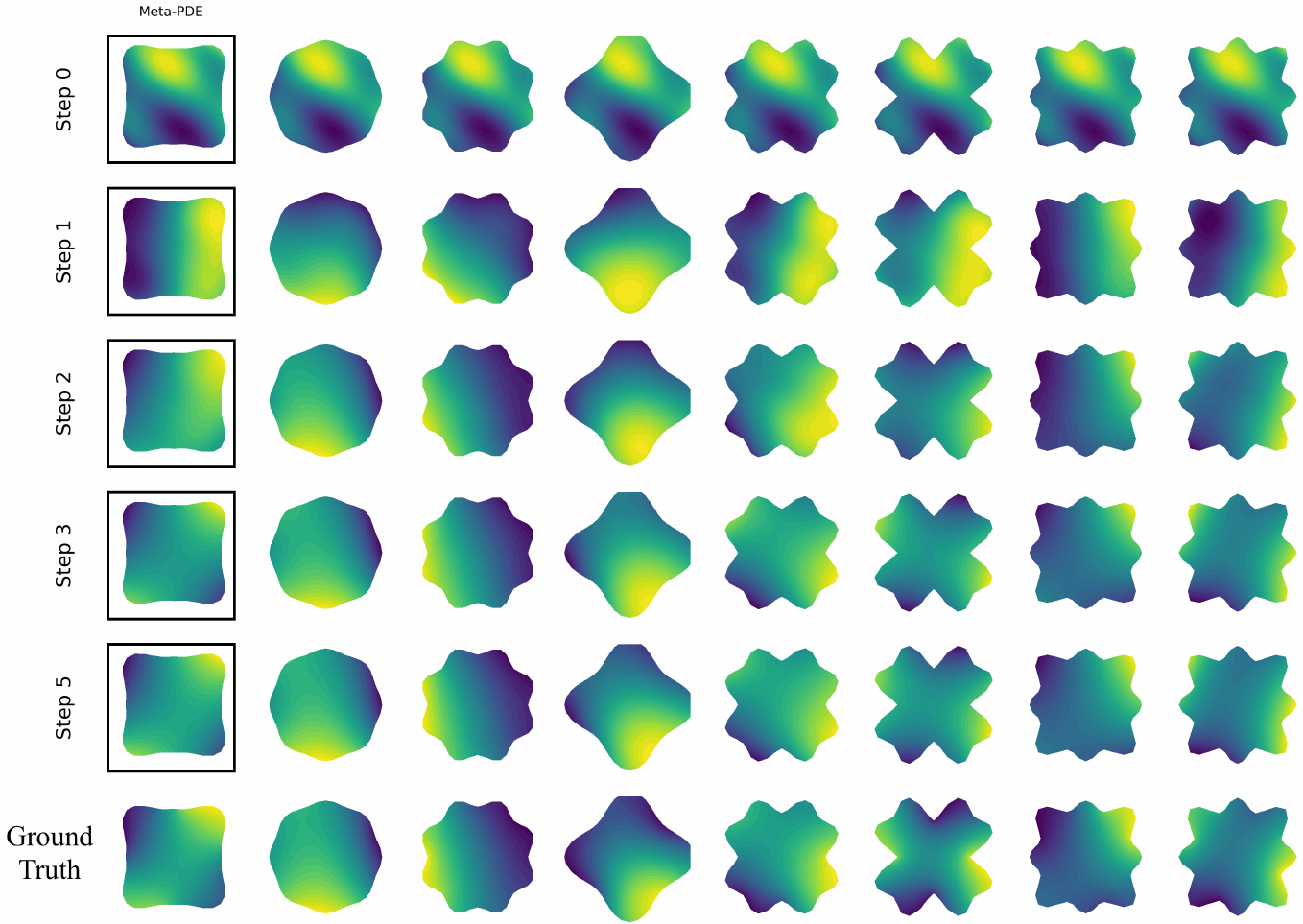}
         \caption{}
         \label{fig:results_per_step}
     \end{subfigure}
     \begin{subfigure}{0.505\textwidth}
         \includegraphics[width=1.1\textwidth]{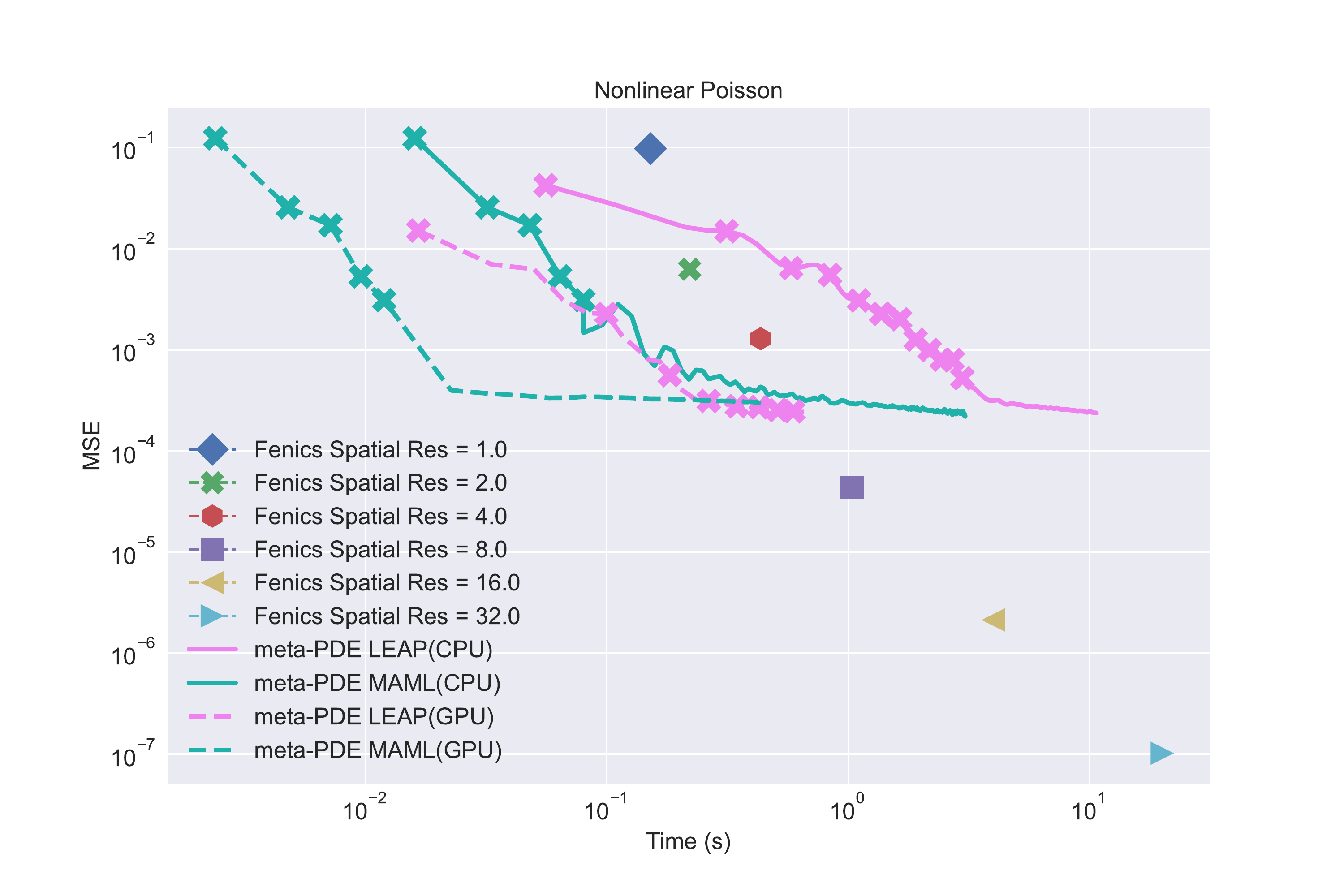}
         \caption{}
         \label{fig:poisson_summary}
     \end{subfigure}
        \caption{(a) Solutions to nonlinear Poisson's equations with varying domains, boundary conditions, and source terms. First Row: Solution represented by Meta-PDE initial NN parameters. Second Row Onwards: Solution after each gradient step in the Meta-PDE inner loop. Bottom: Ground truth FE solution. (b) Speed/Accuracy trade-off for Meta-PDE and FEA. The x-axis is time to solve and y-axis is accuracy, as measured by MSE. For Meta-PDE, we vary the number of training steps after deployment. For FE, we vary the mesh resolutions. Overall, Meta-PDE yields comparable speed/accuracy performance on CPU at intermediate accuracy but better performance on GPU. Meta-PDE reaches an accuracy ceiling at an MSE between $10^{-3}$ and $10^{-4}$, while the FEA solution may be refined to higher accuracy by increasing the resolution of the mesh.}
        \label{fig:poisson}
\end{figure}

Figure \ref{fig:results_per_step} shows the ground truth (baseline) solution for eight PDE problems used in the validation set. The same figure also shows the MAML meta-learned initialization, which can quickly adapt to each PDE problems in five gradient steps.

Figure \ref{fig:poisson_summary} shows the mean squared error (MSE) of each solution and solving time required for the desired accuracy. For the Fenics solution, we vary the mesh resolution (Spatial Res in Figure \ref{fig:poisson_summary}). For the Meta-PDE solution, we increase the number of training steps, starting from the meta-learning initialization. For MAML-based Meta-PDE method, we also start from the meta-learned step size. The highest-resolution FEA method was taken as ground truth and was used to compute MSE. The MSE and solving times were evaluated using 8 held-out problems from the same distribution, and the 8 held-out problems were not used during training. Mean-squared errors are computed between the value of a given approximate solution and the value of the ground truth (highest resolution finite element solution) at 1024 randomly sampled points within the domain. The held-out set configuration remains the same for the other two experiments below. 

Meta-PDE learns to efficiently output moderately accurate solutions. When run on the same CPU (3.6 GHz Intel Xeon Platinum 8000 series) it is about $1-2\times$ faster than a finite element method with similar accuracy. Unlike finite element models, Meta-PDE can be easily accelerated by a GPU, and on GPU we see close to $50\times$ speed up in deployment over similar accuracy CPU-based finite element models.

\subsection{Hyper-Elasticity Equation}
\label{sct:hyperelasticity}
Hyper-elastic materials undergo large shape deformation when force is applied and the stress-strain relation for those materials are highly nonlinear. Rubber is a common example of hyper-elastic materials. the Hyper-elasticity equation models the deformation of those rubber-like materials under different external forces. In particular, we model a homogeneous and isotropic hyper-elastic material under deformation when compressed uniaxially. We assume no additional body or traction force applied to the structure. The goal is to model the final deformation displacement $u$, which maps the material position change from the initial reference position $\mathbf{X}$ to its current deformed location $\bm{x}$.

There are two different approaches to encode the loss function for the hyper-elasticity equation. First, one could directly minimize the residual term in the original strong form, the same approach as we have done for the nonlinear Poisson's equation and for the Burgers' equation. For example, \citet{abueidda2021meshless} has used the first approach to encode the loss function for PINN. Alternatively, one can minimize the Helmholtz free energy of the system and find the corresponding minimizer $u$. We use the second approach to solve the hyper-elasticity equation. See Appendix \ref{appdx:hyperelasticity_pde} for details of the PDE formulation and loss function derivation.

\begin{figure}
     \centering
     \begin{subfigure}[b]{0.485\textwidth}
         \centering
         \includegraphics[width=\textwidth]{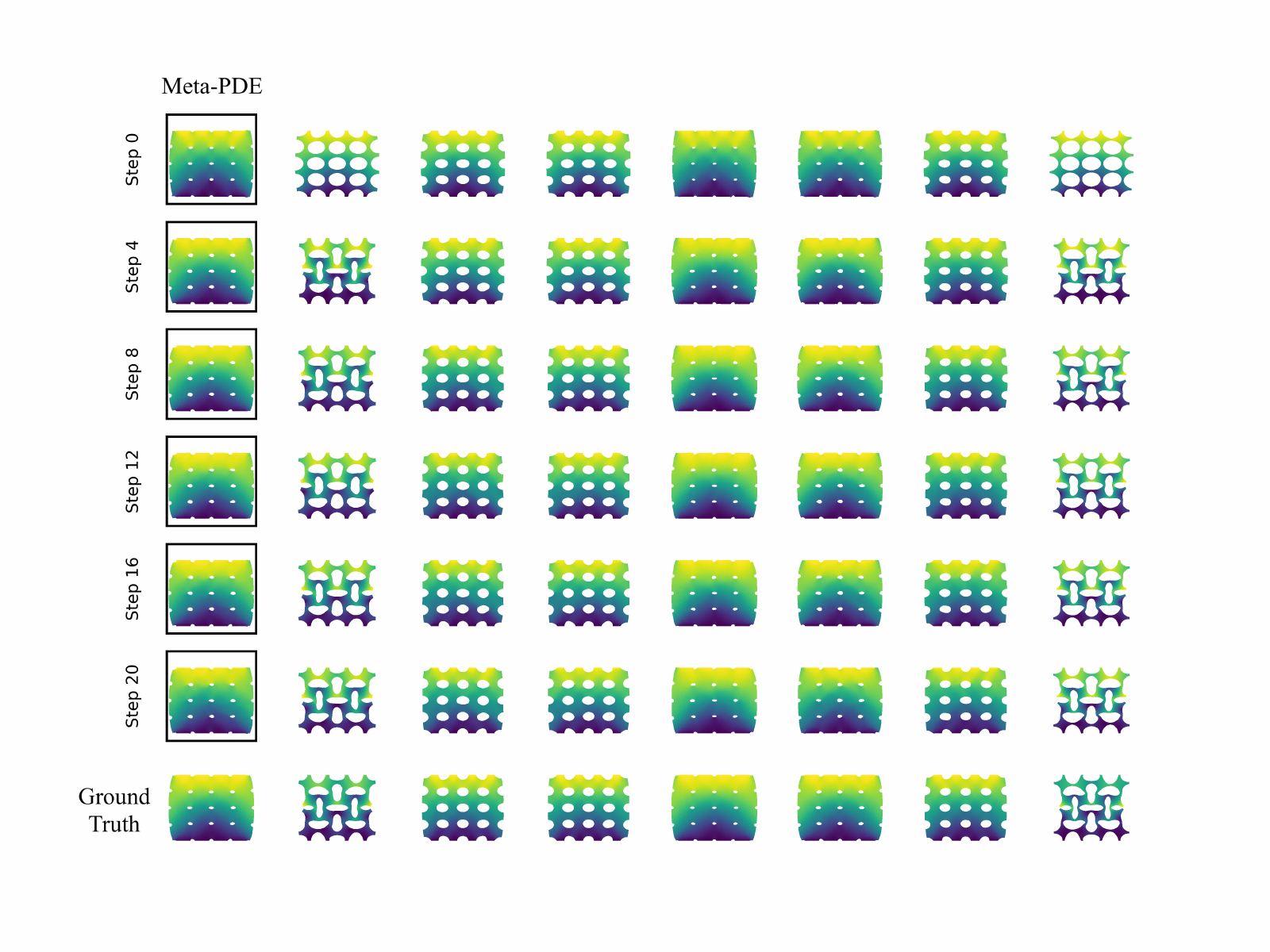}
         \caption{}
         \label{fig:hyperelasticity_per_step}
     \end{subfigure}
     \begin{subfigure}[b]{0.505\textwidth}
         \centering
         \includegraphics[width=0.95\textwidth]{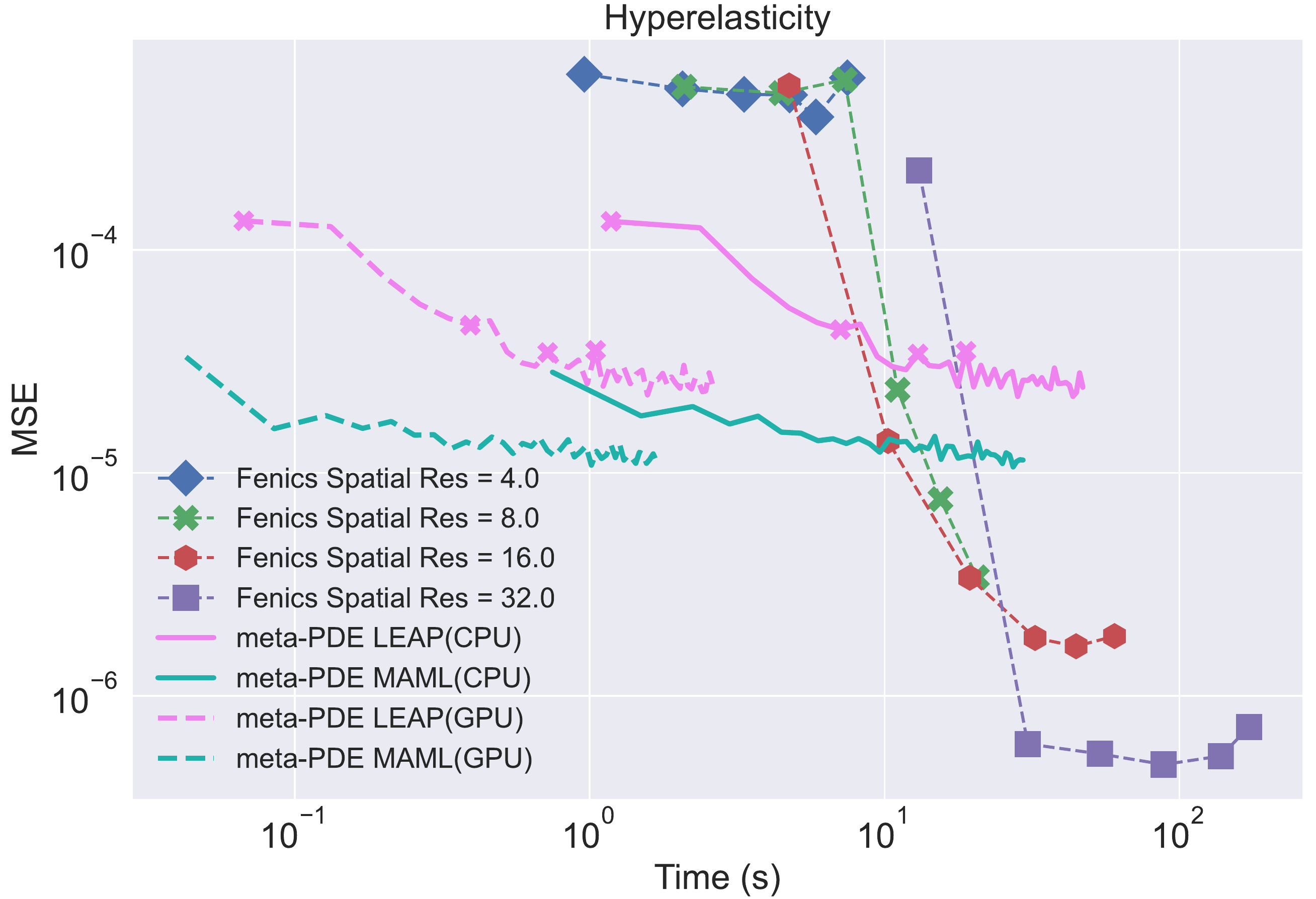}
         \caption{}
         \label{fig:hyperelasricity_summary}
     \end{subfigure}
        \caption{(a) Solutions to hyper-elasticity equations with varying domains. First Row: Solution represented by Meta-PDE initial network parameters. Second Row Onwards: Solution after each gradient step in the Meta-PDE inner loop. Bottom: Ground truth FE solution. (b) Speed/Accuracy trade-off for Meta-PDE and FEA. The x-axis is time to solve and y-axis is accuracy, as measured by MSE. For Meta-PDE, we vary the number of training steps after deployment. For FE, we vary the mesh and boundary resolution. Meta-PDE yields a better speed/accuracy trade-off at intermediate accuracy but is not able to efficiently reach very high accuracy.}
        \label{fig:elasticity}
\end{figure}

We consider the deformation of a two-dimensional porous hyper-elastic material under compression. Their material properties could be very different from their solid counterparts. Because of these interesting differences, the hyper-elastic behavior of porous structures is an active field of research in material science \citep{overvelde2014relating, overvelde2012compaction}. Following the problem setup in \citep{overvelde2014relating}, we use the following parametrized equations to model the shape of the pores:
\begin{align*}
    x_1\! &=\! r(\theta)\cos \theta, \, x_2\! =\! r(\theta)\sin \theta \\
    r(\theta)\! &=\! r_0 \left[ 1\! +\! c_1 \cos(4 \theta)\! +\! c_2\cos(8\theta)\right] \\
    r_0\! &=\! \frac{L_0 \sqrt{2\phi_0}}{\sqrt{\pi \left(2\! +\! x_1^2\! +\! x_2^2\right)}}
\end{align*}
$\phi_0$ is the initial porosity of the structure, and it is sampled uniformly:  $\phi_0 \sim \mathcal{U}(0.0, 0.75)$ in our experiment setup. The parameter pair $(c_1, c_2)$ determines the shape of the pore, and we fixed them to $(0.0, 0.0)$ so that we only work with the circular porous shape. $L_0$ is the initial center-to-center distance between neighboring pores. We also fixed the distance $L_0$ so we work with fixed number of pores on the given material size. With the pore shape and the distance between pore centers $L_0$ fixed, the size of the pore determines the porosity of the structure. The porosity of the structure affects the macroscopic deformation behavior of the structure. 
Figure \ref{fig:hyperelasticity_per_step} shows the ground truth (baseline) solution for eight PDE problems used in the validation set. The same figure also shows the LEAP meta-learned initialization, which can quickly adapt to each PDE problems in 20 gradient steps. 

Figure \ref{fig:hyperelasricity_summary} shows the mean squared error (MSE) of each solution and solving time required for the desired accuracy. Meta-PDE learns to output accurate solutions, and when run on the same CPU is about $5-10\times$ faster than a finite element method with similar accuracy. Running Meta-PDE on a GPU gives close to $100\times$ speed up in deployment over a similar accuracy finite element model run on CPU.

\newpage
\section{Discussion\label{sec:discussion}}
\paragraph{Meta-PDE Methods Comparison}
MAML-based Meta-PDE outperforms LEAP-based Meta-PDE during deployment time in accuracy for a given runtime, while requiring less hyperparameter tuning during training.
We believe that this superior performance is due to MAML having a meta-learned per-state per-parameter step size. 
The advantage of LEAP-based Meta-PDE lies in the meta-training process: LEAP is faster to train and uses less memory than MAML, which required checkpointing for some PDEs.

\begin{figure}
     \centering
     \begin{subfigure}[b]{0.495\textwidth}
         \centering 
         \includegraphics[width=\textwidth]{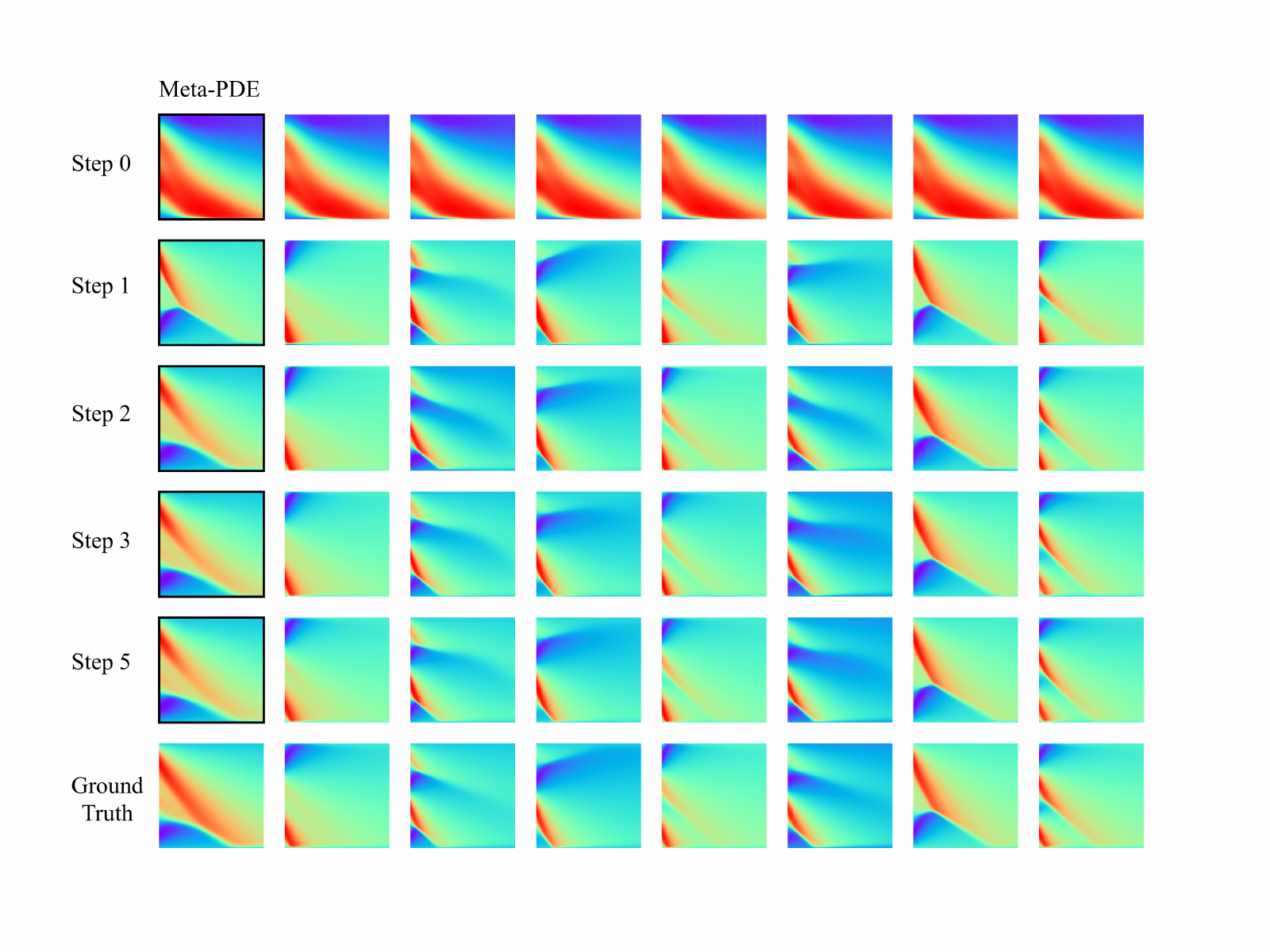}
         \caption{}
         \label{fig:burgers_per_step}
     \end{subfigure}
     \begin{subfigure}[b]{0.495\textwidth}
         \centering
         \includegraphics[width=\textwidth]{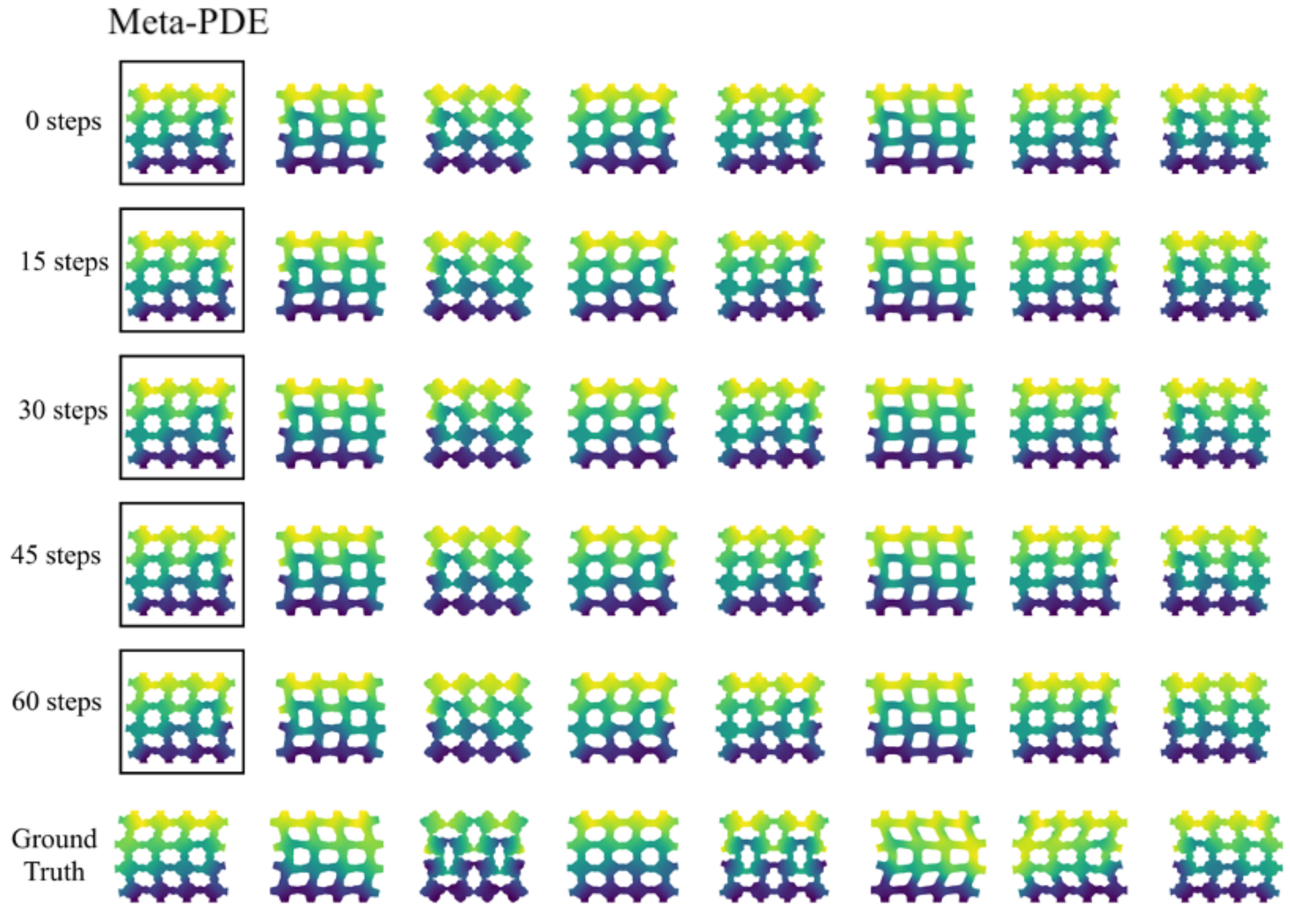}
         \caption{}
         \label{fig:elasticity_flower_meta}
     \end{subfigure}
        \caption{Problems instances where Meta-PDE has difficulty. (a) Solutions to Burgers's Equations with varying initial conditions, and boundary conditions. First Row: Solution represented by Meta-PDE initial NN parameters. Second Row Onwards: Solution after each gradient step in the Meta-PDE inner loop. Bottom: Ground truth FEA solution. Although the Meta-PDE method achieves qualitatively accurate results after a few gradient steps, and is faster than a naive application of FEA, the meta-solver is still slower than a well-chosen baseline for Burgers' equation. (b) Solutions to hyper-elasticity equations with varying porous shape. Top Row: solution represented by Meta-PDE initial neural network parameters. Second row onwards: solution after each gradient step in the Meta-PDE inner loop. Bottom: Ground truth FEA solution. Looking at the bottom two rows, we can see that on many problem instances Meta-PDE fails to find an accurate solution.
        }
        \label{fig:discussionfig}
\end{figure}

\paragraph{Task Domain Generalizability} 
In our study, we restrict each meta-learner to one type of PDE, and for each type of PDE we define the distribution of related tasks via different parameterizations of the same PDE type.
As we increase the volume of this distribution, either by increasing the range of parameters or by increasing the number of parameters, the meta-learning tasks becomes harder. 
As a result, we need to use a larger network architecture, increase the number of inner training steps, and increase the meta-learner's training time to allow it to converge. 
The quality of the final model also deteriorates.
To illustrate this, we looked at a how different pore shapes affect the macroscopic behavior of the material, which is the hyper-elasticity problem studied by \citet{overvelde2014relating}. Following the set up in Section \ref{sct:hyperelasticity}, we fix the initial porosity of the structure $\phi_0 = 0.5$. We now vary the pore shape by sampling $c_1$ and $c_2$ in the parameter pair $(c_1, c_2)$ from a uniform distribution: $c_{1, 2} \sim \mathcal{U}(-0.4, 0.4)$. The initial center-to-center distance between neighboring pore $L_0$ is fixed like before. In this setup, the shape of the pore determines the macroscopic deformation behavior of the structure. As we enlarge the range of possible shapes by increasing the range that we draw possible $(c_1, c_2)$ value from, the accuracy of the solutions produced by Meta-PDE starts to degrade. Figure \ref{fig:elasticity_flower_meta}shows that for many pore shapes, Meta-PDE finds an incorrect solution.

\paragraph{Easy-to-solve PDEs and Hard-to-solve PINNs}  
In addition to our experiments in \cref{sec:experiments} we used Meta-PDE to train a meta-solver for the 1D Burgers' equation. Our distribution of tasks consists of varying initial and boundary conditions. The trained meta-solver was able to consistently achieve accurate results with 5 or fewer gradient steps (see \cref{fig:burgers_per_step}). Although our meta-solver was successful in this regard, and was more efficient than a baseline FEA method written in FEniCS, it was not able to compete with an efficient finite volume method with Godunov flux written in JAX. This suggests that a naive application of Meta-PDE is not well suited for cases where nonlinear FEA under-performs compared to another simple baseline. Appendix \ref{appdx:burgers} and \cref{tbl:hparams,tbl:training} contain a full explanation of the distribution of problems and training hyperparameters for the Burgers' equation experiment. 

We also tried applying Meta-PDE to the Navier-Stokes equations and the 2D Burger's equation. Here, Meta-PDE could not find reasonable solutions for a non-trivial distribution of tasks.
PINNs generally find these time-dependent PDEs hard to solve, so it is not surprising that our meta-learning approach failed to learn how to solve them.
However, recent work \cite{perdikaris2022respectingcausality} has found that by modifying the loss function to better respect the principle of causality in time-dependent PDEs, hard-to-solve PDEs such as the Navier-Stokes equations can be successfully solved with PINNs. 
We are interested in seeing whether incorporating this modified loss function into our meta-learning approach could help us meta-learn solvers for these PDEs.

\section{Conclusion}
Meta-PDE uses meta-learning to amortize PDE solving by accelerating optimization of a PINN representation of the solution. Unlike other fast surrogate models (but like PINNs) our method is mesh-free and data-free, a desirable property when geometry is complex and/or varying across problems within the class we wish to amortize. Unlike PINNs, which are generally too slow to be competitive even with FEA, our method improves on the Pareto frontier of computational cost vs accuracy over FEA. We apply our method to amortize solving of PDEs with varying and complex geometries and terms: non-linear Poisson's equations, hyper-elasticity equations and a 1D Burgers equation. After meta-training, our method both (a) achieves qualitatively correct results for most problems in the distribution and (b) achieves these results after only a few gradient steps, resulting in a solver that is between 1 and 10 times faster than our FEniCS baseline.

This method has some caveats. First, our meta-solvers take a long time to train---several hours on a GPU. Second, our meta-solvers do not have the convergence guarantees that come with methods such as FEA. Third, meta-PDE appears to be better suited for time-independent (i.e., elliptic) PDEs, rather than time-dependent (i.e., hyperbolic) PDEs where information travels along characteristics. For example, we apply Meta-PDE to the 1D Burger's equation, and although we achieve qualitatively accurate results in a few gradient steps, our meta-solver is slower than a strong JAX baseline using the finite volume method with Godunov flux. Finally, there is a vast world of difficult-to-solve PDEs which require specially tailored computational tricks to solve (whether with PINNs or FEA) due to structure in the governing equations, and in this paper we consider only some relatively simple example PDEs. Despite these caveats, we believe that meta-PDE provides a generic and compelling approach to accelerated solving of PDEs with challenging domain geometries without ground-truth data or mesh.

\section{Acknowledgements}
This work was funded by NSF projects IIS-2007278 and OAC2118201, and by the DOE DATASOPT project, DE-AC02-09CH11466.

\bibliographystyle{plainnat}
\bibliography{references}

\newpage
\appendix
\section{MAML Based Meta-PDE}
\label{appdx:meta-pde-dea}

The MAML-based Meta-PDE method learns the model initialization $\theta^0$ for a neural network $f_\theta$ and also learns the learning rate $\alpha$ for each parameter at each inner step. To learn $\theta^0$ we start with a distribution of tasks, where each task represents a different parameterization of the PDE. 
Then we draw a batch of $n$ tasks with individual loss functions $L_i, i \in [n]$. The initialization for each inner task is $\theta^0$.
The gradient update rule for the inner task is simply SGD:
\begin{align*}
    \theta_i^j =  \theta_i^{j-1} - \alpha \nabla_\theta \hat{\mathcal{L}_i}(f_\theta) \quad j \in \lbrace 1, 2, 3, ..., K\rbrace
\end{align*}
We unroll the inner learning loop $K$ steps to find $f_{\theta_i^K}$: the approximate solution for each task $i$ in the batch. 
The meta-loss is the average loss for those $n$ tasks:
\begin{align*}
    \mathcal{L}_\text{MAML} = \frac{1}{n} \sum_{i=1}^n  \hat{\mathcal{L}}_i(f_{\theta^K _i})
\end{align*}
We perform backpropagation through the inner loop to find the gradients w.r.t meta-initialization $\theta^0$ and use the gradients to update $\theta^0$ in the outer loop training:
\begin{align}
    \theta^0 \leftarrow \theta^0 - \beta \frac{1}{n} \nabla_{\theta^0} \sum_{i=1}^n  \hat{\mathcal{L}}_i(f_{\theta^K _i})
\label{eq:maml-update-a}
\end{align}
We also perform backpropagation through the inner loop to find the gradients w.r.t. the per-step, per-parameter step size $\alpha$ and use the gradients to update the $\alpha$ in the outer loop training:
\begin{align}
    {\alpha}  \leftarrow {\alpha} - \beta  \frac{1}{n} \nabla_{\alpha} \sum_{i=1}^n   \hat{\mathcal{L}}_i(f_{\theta^K_i})
\label{eq:maml-update-b}
\end{align}
Eqn. \ref{eq:leap-update} defines the meta-gradient for the LEAP-based Meta-PDE method.
Eqn. \ref{eq:maml-update-a}, \ref{eq:maml-update-b} defines the meta-gradient for the MAML-based Met-PDE method.

\section{Hyper-Elasticity Equation Details}
\label{appdx:hyperelasticity_pde}
The material's initial reference position is $\mathbf{X}$ and its current deformed location is $\bm{x}$. The unknown function $u$ maps the material's position change from the initial reference position $\mathbf{X}$ to its current deformed location $\bm{x}$:
\begin{align*}
    u = \bm{x} - \mathbf{X}
\end{align*}

The deformation gradient $F$ is defined as 
\begin{align*}
    F \equiv \frac{\partial \bm{x}}{\partial \mathbf{X}} = \frac{\partial}{\partial \mathbf{X}} \left(\mathbf{X} + u\right) = \frac{\partial \mathbf{X}}{\partial \mathbf{X}} + \frac{\partial u}{\partial \mathbf{X}} = \mathbf{I} + \frac{\partial u}{\partial \mathbf{X}}
\end{align*}
The constitutive law in continuum mechanics relates Piola-Kirchhoff stress $P$ with deformation gradient $F$ using the following relations:
\begin{align*}
    P = \frac{\partial \psi}{\partial F},
\end{align*}
where $\psi$ is the Helmholtz free energy. For a Neo-Hookean hyperelastic material in 2-D, the energy is given by:
\begin{align*}
    \psi = \frac{1}{2}\lambda \left(\log(J)\right)^2 - \mu \log(J) + \frac{\mu}{2} (\mathbf{I}_c - 2).
\end{align*}
There are two invariants in the above equation. The first is $\mathbf{I}_c \equiv \text{tr}(C) $ and $C$ is the right Cauchy-Green tensor, defined as $C = F^T F$. The second invariant is  $J \equiv \det (F)$. Substitute the two invariant into the above equation, the Piola-Kirchohoff stress $P$ becomes: 
\begin{align*}
    P = \frac{\partial \psi(F)}{\partial F} = \mu F \left(\lambda \ln (J) - \mu\right) F^{-T}
\end{align*}
In the absence of body and traction forces, the Hyperelasticity equations can be written as
\begin{align*}
    \nabla_{\mathbf{X}} \cdot P &= 0 \quad \textnormal{  \hspace{1.0cm}   } \mathbf{X} \in \Omega \\
    \hat{u} &= g(u) \quad \textnormal{  \hspace{0.5cm}   } \mathbf{X} \in \Gamma_u \\
    P \cdot N &= T \quad \textnormal{  \hspace{0.9cm}   } \mathbf{X} \in \Gamma_T. 
\end{align*}
$N$ is the normal vector relative to $\mathbf{X}$. Because the traction force is absent, we set  $T = 0$ on boundary $\Gamma_T$. Dirichlet boundary conditions are imposed on boudnary $\Gamma_u$.\\
The solution $u$ to the Hyperelasticity equations also acts as the minimizer of the Helmholtz free energy $\Pi$ of the entire system:
 \begin{align*}
    u = \text{argmin}_u\ \Pi(u) 
    = \text{argmin}_u\  \left[\int_{\Omega} \psi \, d\bm{x} \right]
\end{align*}
During training, we randomly and uniformly sample collocation points from the PDE domain~$\Omega$ and Dirichlet boundary $\Gamma_u$ and use these points $\mathcal{C} \in \Omega$ and~${\partial \mathcal{C} \in \Gamma_u}$ to form a Monte Carlo estimate of the true loss. For the Hyperelasticity equations, the training loss is
\begin{align*}
    \mathcal{L}_{\text{PINN}} (f_\theta) =\frac{1}{|\mathcal{C}|} \sum_{\bm{x}\in\mathcal{C}} \psi(f_\theta) 
    + \frac{1}{|\partial \mathcal{C}|} \sum_{\bm{x}\in\partial\mathcal{C}} \left|\left| \hat{u} - g(f_\theta) \right|\right| _2 ^2.
\end{align*}

\section{Burger's Equation}
\label{appdx:burgers}
Burger's equation is a time-dependent PDE that models a system consisting of a moving viscous fluid. The 1D version of the equation models the fluid flow through an ideal thin pipe. The strong form of Burger's equation is given by
\begin{align*}
    \frac{\partial u}{\partial t} + u \frac{\partial u}{\partial x} - \nu \frac{\partial ^2 u}{\partial x^2} &= 0, \quad \textnormal{  \hspace{1.0cm}   } x \in \Omega, \textnormal{  \hspace{1.0cm}   } t \in [0, T] \\
    u(x, 0) &= u_0(x), \quad \textnormal{  \hspace{0.3cm}   } x \in \Omega \\
    u(x, t) &= \bar{u}, \quad \textnormal{  \hspace{0.95cm}   } x \in \partial \Omega, \textnormal{  \hspace{0.8cm}   } t \in (0, T] \,.
\end{align*}
The unknown $u(x, t)$ is the speed of the fluid.
If the viscosity~$\nu$ is low, the fluid develops a shock wave.
Following the derivation in Eqn.~\ref{eq:pinn_training_loss_time}, the equation is constrained in the domain with the operator $\mathcal{F} = u \cdot \nicefrac{\partial u} {\partial x} - \nu \cdot \nicefrac{\partial ^2 u}{\partial x^2}$, and constrained on the domain boundary with the operator $\mathcal{G} = u(x, 0) - u_0(x)$. 

\begin{figure}
     \centering
     \begin{subfigure}[b]{0.495\textwidth}
         \centering
         \includegraphics[width=\textwidth]{figures/burgers_meta.pdf}
         \caption{}
         \label{fig:burgers_per_step_appendix}
     \end{subfigure}
     \begin{subfigure}[b]{0.495\textwidth}
         \centering
         \includegraphics[width=\textwidth]{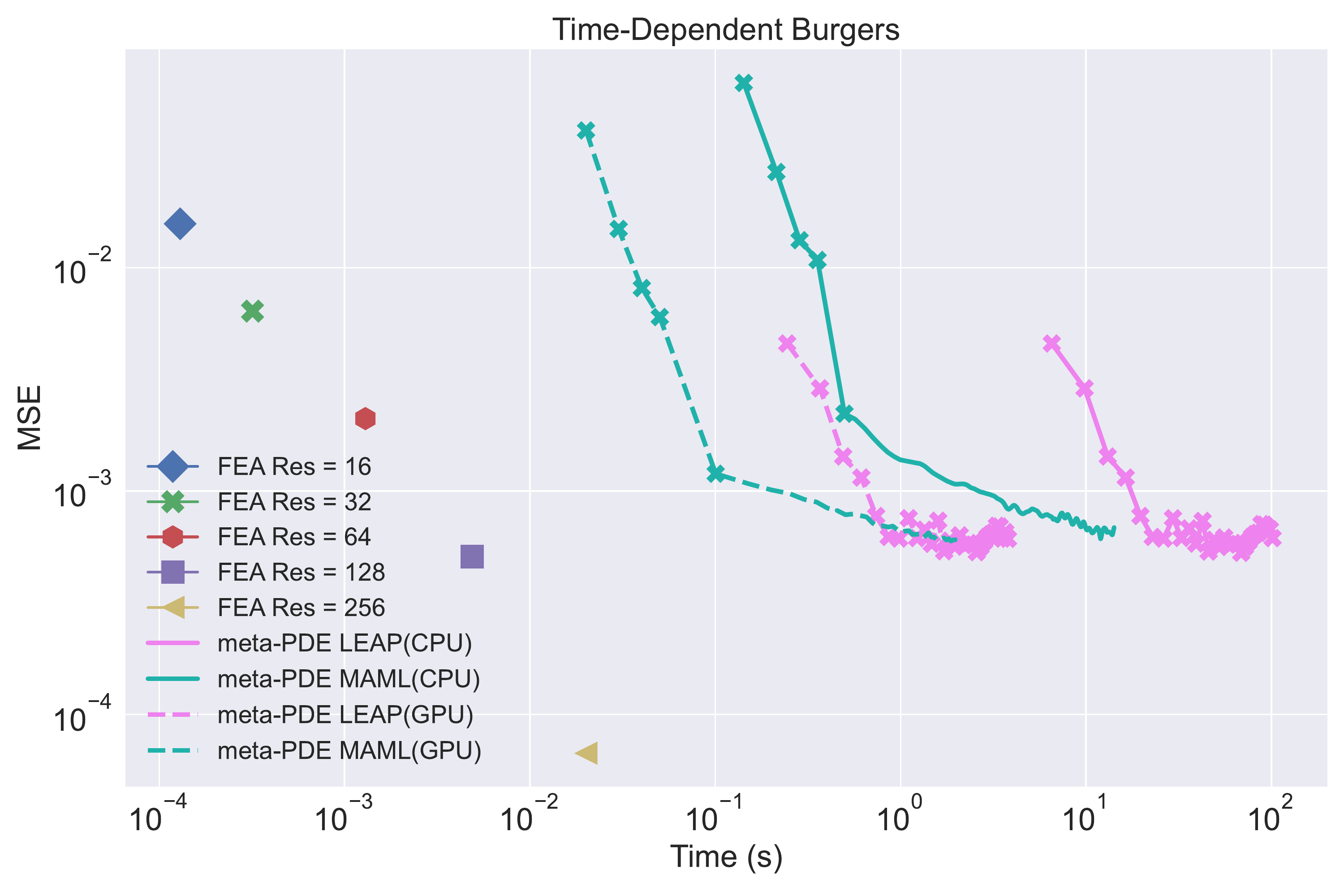}
         \caption{}
         \label{fig:burgers_summary}
     \end{subfigure}
        \caption{(a) Solutions to Burgers's Equations with varying initial conditions and boundary conditions. First Row: Solution represented by Meta-PDE initial NN parameters. Second Row Onwards: Solution after each gradient step in the Meta-PDE inner loop. Bottom: Ground truth FEA solution (b) Speed/Accuracy trade-off for Meta-PDE and FEA. The x-axis is time to solve and y-axis is accuracy, as measured by MSE. For Meta-PDE, we vary the number of training steps after deployment. For FE, we vary the mesh resolution and number of timesteps. 
        }
        \label{fig:burgers}
\end{figure}

We look at the wave formation inside an ideal thin pipe with unit length: $x \in [0, 1]$. We also constrain the time horizon $t \in [0, 1]$.
The initial condition is defined by three sinusoidal functions $u(x) = \sin (\pi x) + \theta_1\sin(2\pi x) + \theta_2\sin(4\pi x)$. 
The varying parameters are $\theta_1, \theta_2 \sim \mathcal{U} (-2.0, 2.0)$.  
Dirichlet boundary conditions are imposed on both the left boundary $x=0$ and the right boundary $x=1$, and are both set to $\bar{u} = 0$. The viscosity~$\nu$ is set to 0.01. 

For the ground truth comparison, we first used a baseline FEniCS solver with implicit Euler for the time integration. The Meta-PDE method outperformed the Fenics baseline by $10 - 20\times$ in speed when executed on the same CPU. However, we found that the FEniCS baseline solver was significantly slower than a finite volume method with Godunov Flux and explicit RK3 timestepping written in JIT-compiled JAX. Figure \ref{fig:burgers_per_step_appendix} shows the ground truth (baseline) solution for the eight PDE problems used in the validation set. The same figure also shows the MAML meta-learned initialization, which can quickly adapt to each PDE problems in 5 gradient steps. Figure \ref{fig:burgers_summary} shows the speed and accuracy for the eight PDE problems solved by the fast finite volume method. Resolution indicates the mesh resolution for the spatial domin and the temporal resolution is fixed to be 10 times the corresponding mesh resolution. We see that Meta-PDE learns to output accurate solutions, but when compared with the fast finite volume method, the Meta-PDE method is slower than the finite element method with similar accuracy.

\end{document}